\documentclass{article}

     \usepackage[nonatbib, final]{ml4mols_2020}

\usepackage[utf8]{inputenc} 
\usepackage[T1]{fontenc}    
\usepackage{hyperref}       
\usepackage{url}            
\usepackage{booktabs}       
\usepackage{amsfonts}       
\usepackage{nicefrac}       
\usepackage{microtype}      

\usepackage{graphicx}
\usepackage{caption}
\usepackage{subcaption}
\usepackage[style=ieee, 
citestyle=numeric-comp,
sorting=none]{biblatex}
\title{Score-Based Generative Models for Molecule Generation}

\author{%
  Dwaraknath Gnaneshwar \\
  Julia Computing, Inc \\
  \texttt{dwaraknath.gnaneshwar@juliacomputing.com} \\
   \And
   Bharath Ramsundar\\
   Deep Forest Sciences Inc. \\
   \texttt{bharath@deepforestsci.com} \\
   \AND
   Dhairya Gandhi \\
   Julia Computing, Inc \\
   \texttt{dhairya@juliacomputing.com} \\
   \And
   Rachel Kurchin \\
   Department of Mechanical Engineering \\ 
   Carnegie Mellon University \\
   \texttt{rkurchin@cmu.edu} \\
   \And
   Venkatasubramanian Viswanathan \\
   Department of Mechanical Engineering \\
   Carnegie Mellon University \\
   \texttt{venkvis@cmu.edu}
}

\begin{document}

\maketitle

\begin{abstract}
Recent advances in generative models have made exploring design spaces easier for de novo molecule generation. However, popular generative models like GANs and normalizing flows face challenges such as training instabilities due to adversarial training and architectural constraints, respectively. Score-based generative models sidestep these challenges by modelling the gradient of the log probability density using a score function approximation, as opposed to modelling the density function directly, and sampling from it using annealed Langevin Dynamics. We believe that score-based generative models could open up new opportunities in molecule generation due to their architectural flexibility, such as replacing the score function with an SE(3) equivariant model. In this work, we lay the foundations by testing the efficacy of score-based models for molecule generation. We train a Transformer-based score function on Self-Referencing Embedded Strings (SELFIES) representations of 1.5 million samples from the ZINC dataset and use the Moses benchmarking framework to evaluate the generated samples on a suite of metrics. 
\end{abstract}

\section{Introduction}

Recent advances in generative models have enabled the design of novel molecules with targeted properties. Prior work has explored Generative Adversarial Networks (GANs) \cite{goodfellow2020generative}, variational autoencoders (VAEs) \cite{kingma2013auto,jin2018junction} and normalizing flows (NFs) \cite{satorras2021n,freyflow,madhawa2019graphnvp} for molecule generation. There are a few significant challenges with each of these approaches. Likelihood-based models language models and NFs place restrictions on the model architecture to ensure a tractable normalizing constant for log-likelihood computation. Such architectural constraints limit model expressivity and increase the complexity of both the training procedure and the model design. Moreover, altering these models to include recent advances requires significant efforts in redesigning to accommodate the constraints.
GANs side-step these training challenges by modelling the probability distribution implicitly. However, their training procedure is prone to instabilities \cite{durall2020combating,thanh2020catastrophic}. 

Recently, score-based generative models \cite{song2019generative} have garnered increasing interest due to their success in generating data of various modalities such as images \cite{song2020improved,dhariwal2021diffusion}, audio \cite{chen2020wavegrad,kong2020diffwave}, and graphs \cite{niu2020permutation}. Score-based models model the gradient of the log probability density function, a quantity known as \textit{score} \cite{liu2016kernelized}, using a model called the \textit{score function}. They do not require having a tractable normalizing constant as they don't use the density function directly in their formulation. An extensive review of score-based models can be found in \ref{score_based_gen_models}, but the key attribute for our purposes is the flexibility of architecture of the score function.

\subsection{Why Score-Based Models?}

Having no restrictions on the score function opens up interesting new avenues in molecule generation that are more challenging with other generative models. For example, consider equivariance under symmetry groups, a desirable property to increase data efficiency and reduce model size without sacrificing expressivity. Reference~\cite{satorras2021n} explores a NF model that is equivariant under Euclidean symmetries to generate 3D molecules. They parameterize a continuous-time flow, where the first derivative is modelled by an Equivariant Graph Neural Network (EGNN) \cite{satorras2021n}. Implementing novel equivariant architectures or other improvements to generative models like GANs or NFs might require significant efforts due to normalization constraints. However, given the fact that score-based models do not place any such restrictions on the architecture of the score function, leveraging new developments is a straightforward matter of drop-in replacement. 

We believe that we can leverage this flexibility to enable novel and high-performance generative models for molecules. However, score-based generative models are not easy to train for molecule generation and it is not yet clear what best practices are to train state-of-the-art score-based models for molecules. In this work, we lay the foundations to establish such best practices, and hope it can serve to ignite a broader discussion within the community. As an initial effort to this end, we present results from training a Transformer-based score function on the ZINC dataset \cite{irwin2012zinc}. 

\section{Score-Based Generative Models}
\label{score_based_gen_models}

Consider a set of samples $\{\mathbf{x}_{i}\}_{i=1}^{N}$ from a data-generating distribution $p_{d}(\mathbf{x})$. Score-Based generative models model the \textit{score} of the probability density $p_{d}(\mathbf{x})$, defined as $\nabla_{\mathbf{x}} \log p_{d}(\mathbf{x})$, the gradient of the log probability density at the input data point. Score is a vector field which points in the direction of greatest growth in log density. The procedure is to generate a starting point and move it towards high density regions along the direction of this vector field. We train a model (typically a neural network) to serve as this \textit{score function} $\mathbf{s}_{\theta}: \mathbb{R}^{D} \rightarrow \mathbb{R}^{D} $, to learn the vector field from a data point using denoising score matching~\cite{hyvarinen2005estimation} such that $\mathbf{s}_{\theta}(\mathbf{x}) \approx \nabla_{\mathbf{x}} \log p_{d}(\mathbf{x})$. 

\paragraph{Score matching} We train the score function on i.i.d. samples from the data distribution with the goal of directly estimating the score without estimating the density function $p_{d}(\mathbf{x})$. However, na\"ive implementations of such score matching can lead to biased results because the loss function is weighted by the $p_d(\mathbf{x})$, resulting in inaccurate estimates of the score in low-density regions. This challenge is addressed in \cite{song2019generative} by training the score functions on perturbed data points (that is, Gaussian noise is added to data and the model is trained to denoise the sample). They observe that perturbing data with random Gaussian noise makes the data distribution more amenable to score-based generative modelling. The overall training procedure then becomes 1) perturbing the data using various levels of noise; and 2) simultaneously estimating scores corresponding to all noise levels by training a single conditional score network. Reference~\cite{song2020improved} proposes using a geometric series of increasing standard deviations $\sigma_{1}<\sigma_{2}..<\sigma_{L}$ to obtain the noise-perturbed distribution $p_{\sigma_{i}}(\mathbf{x})$. The training objective then becomes 
\begin{equation}
    \label{multiple noise scales loss function}
    \sum_{i=1}^{L}\lambda(i) \mathbb{E}_{p_{\sigma_{i}}(\mathbf{x})} \left[ || \mathbf{s}_{\theta}(\mathbf{x}, i) - \nabla_{\mathbf{x}}\log p_{\sigma_{i}}(\mathbf{x}) ||_{2}^{2} \right],
\end{equation}
where $\lambda(i) \in \mathbb{R}_{>0}$ is a positive weighting function. Reference~\cite{song2020improved} also provides practical recommendations to choose the various hyperparameters such as $\sigma_{L}, \sigma_{1}$, $L$, etc.

\paragraph{Langevin Dynamics} We now elaborate the iterative procedure used to draw samples from the trained score-based models. Given a step size $\alpha>0$ and a total number of iterations $T$, an initial sample is drawn from a prior distribution $p_{\pi}(\mathbf{x})$. We iteratively move the sample along the vector field provided by the score function for $T$ steps (we are following standard terminology in this field, but note that the update equation doesn't directly match the standard physics definition of Langevin dynamics): 

\begin{equation}
    \label{Langevin dynamics equation}
    \mathbf{x}_{t} \leftarrow \mathbf{x}_{t-1} + \alpha \nabla_{\mathbf{x}} \log p(\mathbf{x}_{t-1}) + \sqrt{2\alpha} \mathbf{z}_{t}, \quad 1 \leq t \leq T,
\end{equation}

where $\mathbf{z}_{i} \sim \mathcal{N}(0, \mathit{I})$. When $\alpha \rightarrow 0$ and $T \rightarrow \infty $, the $\mathbf{x}_{T}$ obtained converge to a sample from $p_{d}(\mathbf{x})$. Langevin Dynamics does not access $p_d(\mathbf{x})$ directly, instead we use $\nabla_{\mathbf{x}} \log p(\mathbf{x})$ through the score function $\mathbf{s}_{\theta}(\mathbf{x})$. Reference~\cite{song2019generative} incorporates the multiple noise levels solution to address the challenge of inaccurate score estimates by running Langevin Dynamics sequentially on the noise levels. This method is called annealed Langevin Dynamics. 

\begin{figure}[t]
    \centering
    \includegraphics[width=10cm]{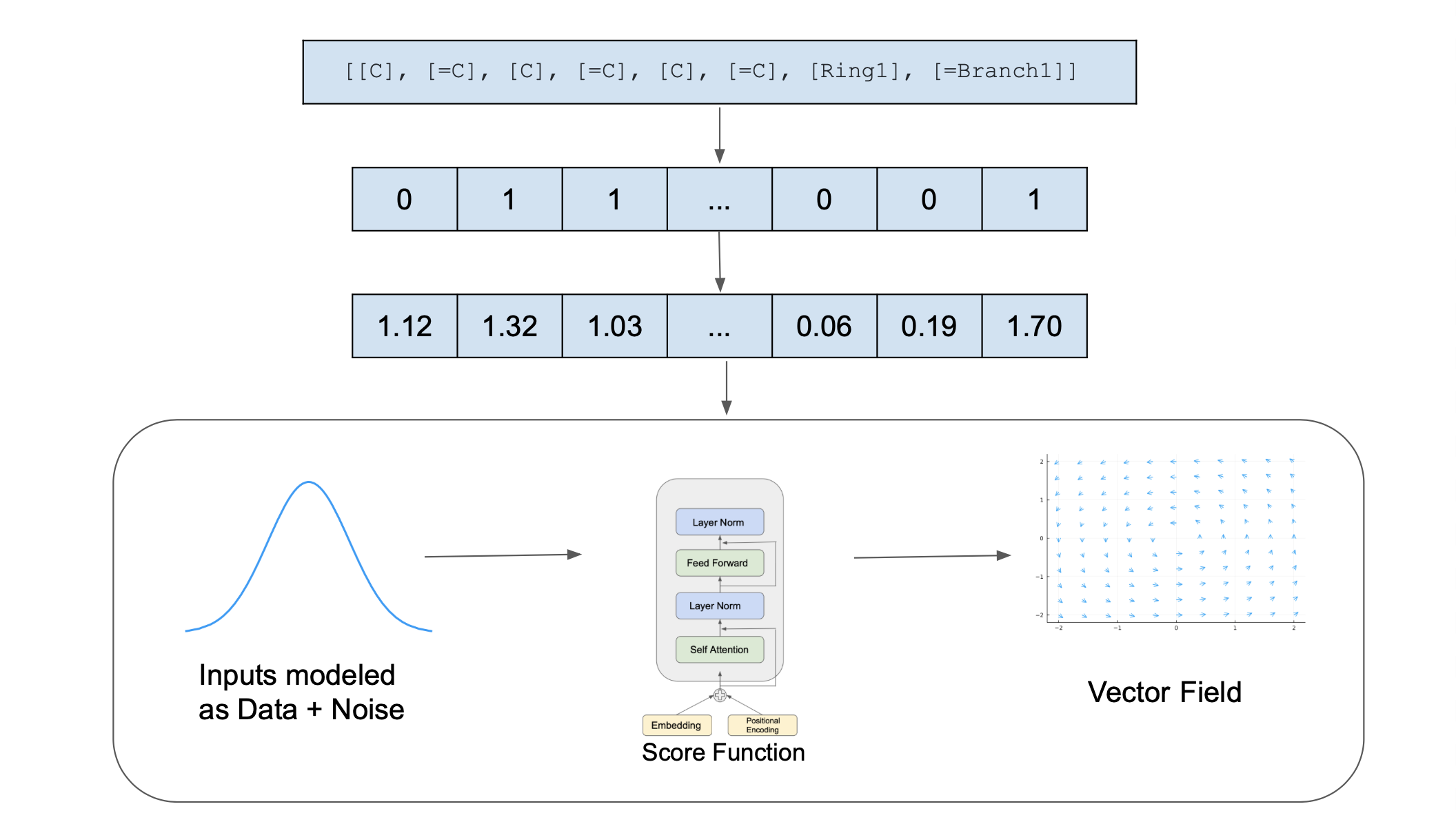}
    \caption{The training flow for our score-based architecture. Compounds are input as SELFIES strings, which are converted into a one-hot encoding. Small amounts of Gaussian noise are added to this encoding, and is then passed into a transformer score function which outputs the vector field.}
    \label{fig:architecture}
\end{figure}

\section{Experiments}
\label{experiments}

We test the capacity of score-based generative models to generate molecules from their text representations in a canonical setting. We train our models on 1.6M drug-like molecules from the ZINC dataset \cite{irwin2012zinc} for 150k steps with a batch size of 256. We use a Transformer model as our score function. The molecules are represented using the SELF-Referencing Embedded Strings (SELFIEs) representation \cite{krenn2020self}. We use the MOSES framework \cite{polykovskiy2018molecular} to evaluate our generated samples. All metrics are calculated on a set of 30k newly generated molecules from our model, following the recommendation from the MOSES framework. We trained all our models on a single Nvidia Tesla V100 GPU. Training the model took approximately 12 hours while time taken for sampling was largely dependent on the number of time steps for each noise scale ($T$). With $T=10$, generating 30k samples took approximately 4 hours.

\paragraph{Hyperparameters} The score functions are modelled after the encoder layers of Transformer \cite{vaswani2017attention} with six layers. The hidden dimensions of the layers are 512 with four attention heads. We use the Adam optimizer \cite{kingma2014adam} with learning rate $10^{-4}$. Following the recommendations made in \cite{song2020improved}, we set $\sigma_{1}$ to 10, which is the maximum Euclidean distance between all pairs of training data points. (We encode the SELFIES string for a sample as a $V\times L$ matrix. Here $V$ is the vocabulary size for tokens, and $L$ is the maximum string length in the dataset. Each column is a one-hot encoded representation of the token at that position. We flatten this binary matrix into a vector and compute the $L^2$ distance). We chose $\sigma_{1} = 0.01$ and a total of 350 levels between $\sigma_{L}, \sigma_{1}$. 
 
\begin{table}[h!]
\centering
\begin{tabular}{ |c|c|c|c|c| } 
 \hline
 \textbf{Metric} & \textbf{HMM} & \textbf{VAE} & \textbf{JTN-VAE} & \textbf{This Work} \\
 \hline 
 \textbf{Valid} ($\uparrow$) & 0.07 & 0.97 & 1.0 & $\textbf{1.0} \pm 0.0$ \\ 
 \textbf{Unique@1k} ($\uparrow$) & 0.62 & 1.0 & 1.0 & $0.88 \pm 0.06$ \\ 
 \textbf{Unique@10k} ($\uparrow$) & 0.56 & 0.99 & 0.99 & $0.82 \pm 0.05$ \\ 
 \textbf{Filters} ($\uparrow$) & 0.90 & 0.99 & 0.97 & $0.37 \pm 0.11$ \\ 
 \textbf{Novelty} ($\uparrow$) & 0.99 & 0.69 & 0.91 & $\textbf{1.0} \pm 0.0$ \\ 
 
 $\textbf{IntDiv}_{1}$ & 0.84 & 0.85 & 0.85 & $\textbf{0.90} \pm 0.04$ \\
 $\textbf{IntDiv}_{2}$ & 0.81 & 0.85 & 0.84 & $\textbf{0.88} \pm 0.04$ \\
 $\textbf{FCD/Test} (\downarrow)$ & 24.46 & 0.09 & 0.39 & $39.84 \pm 2.63$ \\
 $\textbf{FCD/TestSF} (\downarrow)$ & 25.43 & 0.56 & 0.93 & $40.92 \pm 2.77$ \\ 
 \hline
\end{tabular}
\vspace{5pt}
\caption{Experiment results on the 1.6M sample version of Zinc dataset with a Tranformer model as the score function. All results are reported over three independent model initializations. Generated samples are novel and diverse, but don't closely match the training distribution. $\uparrow$ and $\downarrow$ mean higher and lower are better.}`
\label{table:1}
\end{table}

\begin{figure}[t]
    \centering
    \includegraphics[width=8cm]{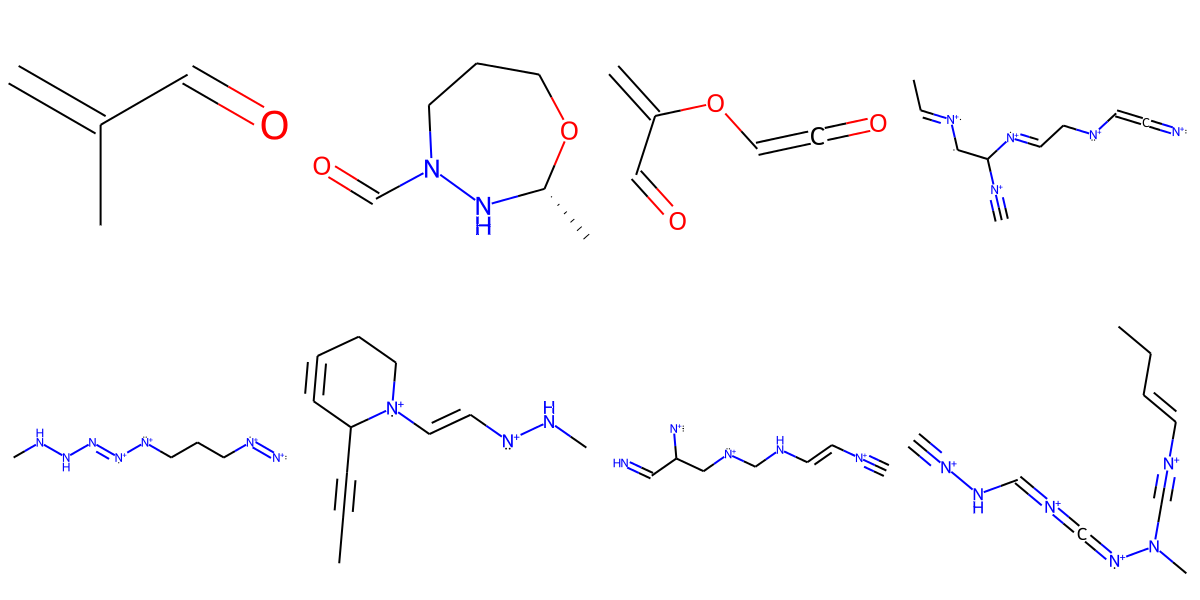}
    \caption{Randomly sampled molecules from a set of 30k molecules generated from score-based model.}
    \label{fig:architecture}
\end{figure}

\section{Discussion}
\label{discussion}

Results from running MOSES benchmarks are shown in Table~\ref{table:1}. High novelty and validity percentages indicate that the model is able to generate a novel set of structurally valid compounds. Uniqueness scores are lower than VAE models but are still competitive. Uniqueness could possibly be improved with finer tuning of the noise scales. High $\textbf{IntDiv}_{1}, \textbf{IntDiv}_{2}$ scores, which assess the chemical diversity within the generated set, suggest that the model is not suffering from mode collapse, as mode collapse often leads to producing limited variety of samples. Low \textbf{Filters} scores and \textbf{FCD} scores indicate that while the generated molecules are chemically valid, they are far from the training distribution. On the one hand, generating diverse chemical matter distinct from the training distribution can prove valuable, but it is possible the model lacks the representation power to fully model the training distribution.

\section{Acknowledgements}
The information, data, or work presented herein was funded in part by the Advanced Research Projects Agency-Energy (ARPA-E), U.S. Department of Energy, under Award Number DE-AR0001211. The views and opinions of authors expressed herein do not necessarily state or reflect those of the United States Government or any agency thereof.

\section{Conclusion and Future Work}

In this work, we explore the use of score-based generative models for the task of molecule generation. We train a Transformer-based score-based generative model on the 1.6 million sample version of the ZINC dataset and evaluate the generated molecules using the MOSES framework. The results indicate that score-based models are capable of generating diverse and unique molecules but the generated points, however, are far from the training distribution. 

Future work will improve the score function models to include structural information about the molecule and train score-based models to generate molecule graphs directly instead of SELFIES. We plan to incorporate recent advances in equivariant operations to test the efficacy of equivariant score-based generative models for molecule graph generation. We hypothesize that richer choices of models may enable score-based methods to fully learn rich training distributions. Alternatively, it is possible that training a large enough transformer model may suffice to learn these distributions. We plan to explore both hypotheses.

\printbibliography
\end{document}